\pdfoutput=1

\documentclass[11pt]{article}

\usepackage{acl}

\usepackage{times}
\usepackage{latexsym}
\usepackage{enumitem}
\setlist{nosep}
\usepackage{graphicx}
\usepackage{multirow}
\usepackage{caption}
\usepackage{subcaption}
\usepackage{soul}
\usepackage{float}
\setul{1pt}{.6pt}
\usepackage[T1]{fontenc}

\usepackage[utf8]{inputenc}

\usepackage{microtype}

\title{EnCBP: A New Benchmark Dataset for \\Finer-Grained Cultural Background Prediction in English}

\author{Weicheng Ma, Samiha Datta, Lili Wang, \and Soroush Vosoughi \\
        Department of Computer Science, Dartmouth College \\ 
        \{weicheng.ma.gr, samiha.datta.23, lili.wang.gr, soroush.vosoughi\}@dartmouth.edu}

\begin{document}
\maketitle
\begin{abstract}
While cultural backgrounds have been shown to affect linguistic expressions, existing natural language processing (NLP) research on culture modeling is overly coarse-grained and does not examine cultural differences among speakers of the same language.
To address this problem and augment NLP models with cultural background features, we collect, annotate, manually validate, and benchmark EnCBP, a finer-grained news-based cultural background prediction dataset in English.
Through language modeling (LM) evaluations and manual analyses, we confirm that there are noticeable differences in linguistic expressions among five English-speaking countries and across four states in the US.
Additionally, our evaluations on nine syntactic (CoNLL-2003), semantic (PAWS-Wiki, QNLI, STS-B, and RTE), and psycholinguistic tasks (SST-5, SST-2, Emotion, and Go-Emotions) show that, while introducing cultural background information does not benefit the Go-Emotions task due to text domain conflicts, it noticeably improves deep learning (DL) model performance on other tasks.
Our findings strongly support the importance of cultural background modeling to a wide variety of NLP tasks and demonstrate the applicability of EnCBP in culture-related research.
\end{abstract}

\section{Introduction}
Psychological research has revealed that people from different cultural background behave differently in the ways they think \cite{culture-think-1}, talk \cite{culture-talk-1}, write \cite{culture-writing-style-1,culture-writing-style-2,culture-writing-style-3}, and express emotions \cite{culture-emotion-1,culture-emotion-2,culture-emotion-3}.
NLP researchers have applied cultural background information to model differences in linguistic expressions across culture groups especially for psycholinguistic tasks 
\footnote{In this paper, we refer to NLP tasks reflecting the psychological states of people, e.g., sentiments and emotions, as psycholinguistic tasks.}, e.g., distributional perspective identification \cite{culture-classify-language-1} and sentiment analysis \cite{culture-emotion-2}.
In prior research, culture groups are usually defined by official language \cite{culture-classify-language-1} (e.g., US, UK, and India are considered part of the same culture group) or, even more coarse-grained, by ideology \cite{culture-classify-ideology-1} (e.g., ``Western" countries and ``Eastern" countries).
These settings typically overlook the nuanced cultural differences across or within countries, and they do not provide useful information for modeling different language use behavior in mono-lingual contexts.

To study culture-specific linguistic expressions in the same language and to apply culture-related knowledge to other NLP tasks, we build EnCBP, a cultural background prediction dataset in English.
Following \cite{culture-specific-language-1}, we assume that language use patterns are more consistent inside each country or each district in a large country, e.g., states in the US.
As such, we first construct news corpora by sampling news articles covering five frequently discussed and controversial topics from major news outlets in five English-speaking countries and four geographically dispersed states in the US.
We then break the articles down to paragraphs and annotate them with the country and state codes of the news outlets to construct the country- and district-level subsets of EnCBP.
We refer to the two subsets as EnCBP-country and EnCBP-district.
To ensure annotation quality, we randomly sample 20 instances from each culture group and have them validated manually by local residents using Amazon Mechanical Turk (MTurk).
The annotation accuracies and inter-validator agreement rates are both high for all the validation sets, supporting the correctness of the labels and demonstrating the differences in writing style across culture groups.
In addition, we benchmark EnCBP for cultural background prediction with three widely-used NLP model architectures, namely BiLSTM, BERT \cite{bert-orig}, and RoBERTa \cite{roberta-orig}.
Among the three models, the roberta-base model achieves the best overall performance, scoring 82.96 in F1-macro on the EnCBP-country and 73.96 on EnCBP-district.
The better performance of BERT and RoBERTa over BiLSTM implies the importance of deep neural network architectures and large-scale pre-training for the challenging text-based cultural background prediction task.

We conduct both quantitative and qualitative analyses on EnCBP to show the differences in linguistic expressions across culture groups.
For the quantitative analysis, we fine-tune a BERT model on the corpus with each cultural background label and evaluate it on the corpora of all the culture groups.
Results show that all the fine-tuned models are more compatible with the cultural domains of their training corpora and less compatible with the those of other corpora, with perplexity differences ranging from 0.43 to 14.90.
For the qualitative analysis, we manually analyze sentence structures and the choices of words or phrases in instances randomly sampled from EnCBP to illustrate culture-specific English expressions.

Furthermore, we evaluate a BERT model on nine psycholinguistic (sentiment analysis and emotion recognition), syntactic (named entity recognition), and semantic (paraphrase identification, natural language inference, semantic textual similarity, and text entailment) tasks to examine how modeling culture-specific English writing styles benefits the performance of NLP models.
The models that incorporate cultural background information perform noticeably better on the named entity recognition (NER) task, most semantic tasks, and the sentiment analysis (SA) tasks.
In our emotion recognition (ER) evaluation on the Go-Emotions dataset, however, the performance is slightly harmed by incorporating cultural background information.
This is likely due to the imbalanced cultural background distribution in the dataset, as the evaluation performance of BERT clearly improves on Emotion, another ER dataset.
On the paraphrase identification (PI) task, while the model performs better with cultural information incorporated, the improvement is lower than those on SA and NER tasks.
This result suggests that differentiating linguistic expressions with the same semantic meaning may introduce additional noise to semantic tasks.

Our analyses and evaluations support the importance of cultural background modeling for a wide range of NLP tasks and show that EnCBP can contribute to future culture-related NLP research.

The contributions of this paper are three-fold:
\begin{itemize}[leftmargin=*,topsep=0pt]
    \item we construct, manually validate, and benchmark EnCBP, a mono-lingual news-based cultural background prediction dataset;
    \item we qualitatively and quantitatively examine the distinctions in writing style from different culture groups; and
    \item we show the effect of introducing cultural background information to nine downstream NLP tasks to showcase the importance of cultural information in natural language understanding.
\end{itemize}

\section{Dataset Construction}
This section introduces the construction, validation, and benchmarking of the EnCBP dataset.
The EnCBP dataset adopts a multi-class classification objective.
The labels are country codes of news outlets for the coarse-grained subset (EnCBP-country) and US state codes for the finer-grained subset (EnCBP-district).
\begin{table*}[h]
\centering
\begin{tabular}{|l|c|c|c|c|c|c|c|c|c|c|} 
\hline
\multicolumn{2}{|l|}{\multirow{2}{*}{}} & \multicolumn{5}{c|}{Topics}                                                                                                                                                                                                                                                                                                     & \multicolumn{4}{c|}{Splits}                                                                                     \\ 
\cline{3-11}
\multicolumn{2}{|l|}{}                  & \multicolumn{1}{c|}{\begin{tabular}[c]{@{}c@{}}Global\\ Warming \end{tabular}} & \multicolumn{1}{c|}{Abortion} & \multicolumn{1}{c|}{\begin{tabular}[c]{@{}c@{}}Immi-\\gration\end{tabular}} & \multicolumn{1}{c|}{\begin{tabular}[c]{@{}c@{}}Social\\ Safety\\ Net \end{tabular}} & \multicolumn{1}{c|}{\begin{tabular}[c]{@{}c@{}}Mandatory\\ Vaccination \end{tabular}} & \multicolumn{1}{c|}{Total} & \multicolumn{1}{c|}{Train} & \multicolumn{1}{c|}{Dev} & \multicolumn{1}{c|}{Test}  \\ 
\hline
\multirow{9}{*}{\rotatebox[origin=c]{90}{Labels}} & US            & 332                                                                            & 455                           & 253                              & 336                                                                                 & 624                                                                                   & 2,000                      & 1,600                      & 200                      & 200                        \\ 
\cline{2-11}
                        & UK            & 648                                                                            & 129                           & 383                              & 456                                                                                 & 384                                                                                   & 2,000                      & 1,600                      & 200                      & 200                        \\ 
\cline{2-11}
                        & AUS     & 532                                                                            & 188                           & 439                              & 402                                                                                 & 439                                                                                   & 2,000                      & 1,600                      & 200                      & 200                        \\ 
\cline{2-11}
                        & CAN        & 418                                                                            & 379                           & 430                              & 315                                                                                 & 458                                                                                   & 2,000                      & 1,600                      & 200                      & 200                        \\ 
\cline{2-11}
                        & IND         & 478                                                                            & 171                           & 540                              & 371                                                                                 & 440                                                                                   & 2,000                      & 1,600                      & 200                      & 200                        \\ 
\cline{2-11}
                        & NY            & 206                                                                            & 134                           & 443                              & 704                                                                                 & 513                                                                                   & 2,000                      & 1,600                      & 200                      & 200                        \\ 
\cline{2-11}
                        & CA            & 274                                                                            & 242                           & 473                              & 556                                                                                 & 455                                                                                   & 2,000                      & 1,600                      & 200                      & 200                        \\ 
\cline{2-11}
                        & GA            & 245                                                                            & 384                           & 214                              & 389                                                                                 & 768                                                                                   & 2,000                      & 1,600                      & 200                      & 200                        \\ 
\cline{2-11}
                        & TX            & 365                                                                            & 328                           & 468                              & 585                                                                                 & 254                                                                                   & 2,000                      & 1,600                      & 200                      & 200                        \\
\hline
\end{tabular}
\caption{Number of documents associated with each label and under each topic in EnCBP. For each country or district label, the documents under each topic are randomly sampled into the training, development, and test sets with a 80\%/10\%/10\% split.}
\label{tbl:dataset-statistics}
\end{table*}
\subsection{Data Collection and Annotation}
Our work relies on the hypothesis that news articles from mainstream news outlets of a country or district reflect the local language use patterns.
Thus, we construct 5 text corpora with news articles posted by New York Times, Fox News, and the Wall Street Journal in the US, BBC in UK, Big News Network - Canada in Canada (CAN), Sydney Morning Herald in Australia (AUS), and Times of India in India (IND) for EnCBP-country.
For EnCBP-district, we construct 4 corpora from Coosa Valley News, WJCL, and Macon Daily in Georgia (GA), Times Union, Gotham Gazette, and Newsday in New York (NY), NBC Los Angeles, LA Times, and San Diego Union Tribune in California (CA), and Hardin County News, Jasper Newsboy, and El Paso Times in Texas (TX).
We stream news articles from Media Cloud \footnote{https://mediacloud.org/}, a platform that collects articles from a large number of media outlets, using its official API.

To maintain consistent mentions of events and named entities (NEs) in the corpora, we limit the articles to those under five frequently discussed topics, namely ``global warming", ``abortion", ``immigration", ``social safety net", and ``mandatory vaccination".
1,000 news articles published between Jan. 1, 2020 and Jun. 30, 2021 are sampled from each news outlet to form our corpora.

After data collection, we remove duplicates and overly short documents (less than 100 words) to ensure data quality.
We also replace the mentions of countries and districts with the ``[country]" and ``[district]" special tokens.
Then, we chunk the remaining news articles into paragraphs and label the documents with the country or district codes of the news outlets by which they are posted.
We adopt paragraph-level annotations since asking the validators to read an overly-long document may cause them to lose track of culture-specific information when they are making judgments.
Most state-of-the-art DL models also have input length limits that are not capable of encoding full-length news article.
To avoid overly simplifying the task, we remove paragraphs containing NE mentions that are mainly used by news media in specific countries or districts.
We quantify the specificity of NEs using inverse document frequency (IDF) scores.

From the filtered news paragraphs, we sample 2,000 paragraphs from the corpus of each culture group to form the annotated dataset.
Table \ref{tbl:dataset-statistics} provides the statistics of the label and topic distribution of the instances in EnCBP.


\subsection{Manual Validation}
\begin{table}
\centering
\begin{tabular}{|c|c|c|} 
\hline
\begin{tabular}[c]{@{}c@{}}Culture\\Groups\end{tabular} & \multicolumn{1}{c|}{ACC (\%)} & \multicolumn{1}{c|}{IAA}  \\ 
\hline
US                                                      & 64.00                    & 0.61                                                    \\ 
\hline
UK                                                      & 76.67                    & 0.73                                                    \\ 
\hline
AUS                                                     &74.00&0.71\\ 
\hline
CAN                                                     & 58.67                    & 0.57                                                    \\ 
\hline
IND                                                     & 61.43                    & 0.61                                                    \\ 
\hline
NY                                                      & 81.33                    & 0.78                                                    \\ 
\hline
CA                                                      & 64.67                    & 0.59                                                    \\ 
\hline
GA                                                      & 70.00                    & 0.66                                                    \\ 
\hline
TX                                                      & 72.00                    & 0.68                                                    \\
\hline
\end{tabular}
\caption{Validation results of the EnCBP dataset. ACC and IAA refer to validation accuracy and inter-annotator agreement rate in Fleiss' $\kappa$, respectively.}
\label{tbl:inter-annotator-agreement}
\end{table}
To ensure that the cultural background labels in EnCBP correlate with writing styles, we randomly sample 50 instances from each class and manually validate them on MTurk.
In each questionnaire, we pair the sampled instance with another random news paragraph from EnCBP and ask three annotators whether the first, second, or both paragraphs are posted by media outlets in a specific country or district.
We manually check the instances to ensure there are no country- or district-specific mentions remaining to avoid potential information leakage.
For quality control purposes, we only hire crowdsourcing workers from the country or US state matching the label of the instances sampled for validation.

\begin{figure*}[!h]
    \centering
    \includegraphics[width=1.\linewidth]{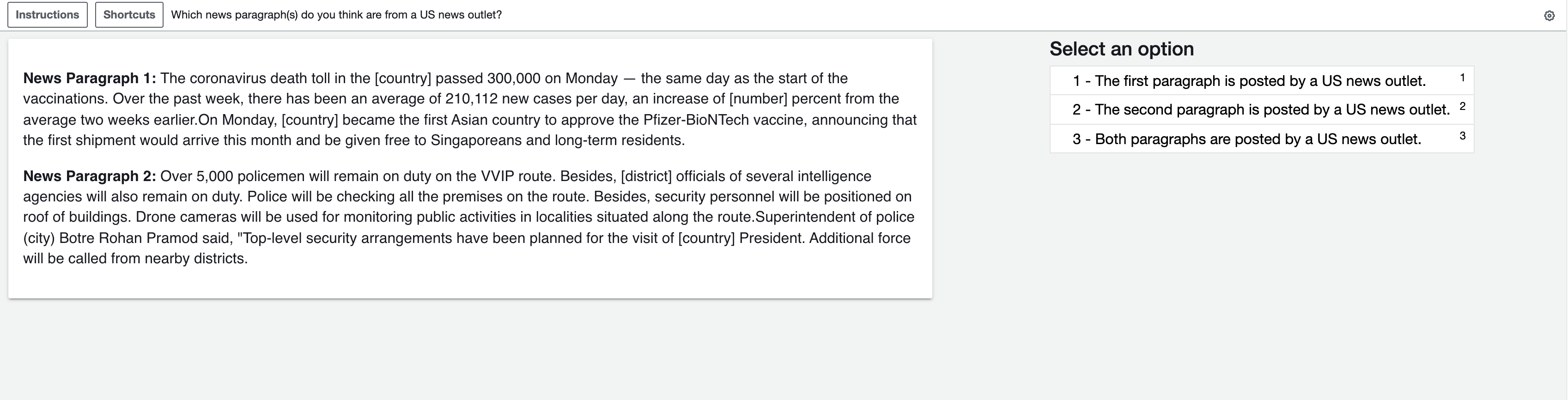}
    \caption{An example of the questionnaire used for validating the annotations in EnCBP.}
    \label{fig:questionnaire-example}
\end{figure*}
To ensure the quality of annotations in EnCBP, we hire crowdsourcing workers from MTurk to validate randomly sampled data points. 
Since all the news articles are written by native English speakers and the culture groups are not strictly separated from each other, it is difficult for a validator to identify whether a news paragraph is written by a journalist from the same cultural background as them.
Instead, we provide each validator with a news paragraph posted by an international or domestic news outlet in the country or district they live in (MTurk allows for filtering based on location) and a randomly selected news paragraph from our dataset.
The validators are asked to compare the two news paragraphs and decide which of the two paragraphs (or both) were written by their local news outlets through analyzing the use of words, phrases, and sentence structures.
To avoid information leak and bias in the validation process, the mentions of countries and districts are replaced with ``[country]" and ``[district]" special tokens at the pre-processing stage of the dataset.
An example questionnaire is shown in Figure \ref{fig:questionnaire-example}.

We display the validation accuracy (ACC), i.e., the proportion of the validators' answers that match the labels of those instances in EnCBP, and inter-annotator agreement rate (IAA) in Table \ref{tbl:inter-annotator-agreement}.
Since we have three options in each of the questionnaires, the ACC of random guess is around 33\% for each culture group.
We quantify IAA with Fleiss' $\kappa$ \cite{fleiss-kappa-orig}, a widely used metric for evaluating IAA.
The Fleiss' $\kappa$ in Table \ref{tbl:inter-annotator-agreement} range from moderate ($>0.40$) to substantial agreement ($>0.60$).
We infer from the relatively high ACC and IAA that: 1) news writing styles are affected by the cultural backgrounds of journalists and 2) writing styles in each culture group are identifiable by local residents.
Since we removed country- or state-specific NEs and mentions of countries or states from the paragraphs, and as the distributions of topics and sentiments are balanced across corpora, the chance that the validators make their judgments based on these external information is low.


\subsection{Dataset Benchmarking}
\begin{table}[]
\centering
\begin{tabular}{lc|c}
\hline
\multicolumn{1}{l}{Model} & \multicolumn{1}{c|}{EnCBP-country} & \multicolumn{1}{c}{EnCBP-district} \\ \hline
BiLSTM                    &50.89 (0.98)&44.53 (1.39)\\ \hline
BERT                      &78.13 (0.67)&72.09 (1.84)\\ \hline
RoBERTa                   &82.96 (0.89)&73.96 (1.01)\\ \hline
\end{tabular}
\caption{Benchmark performance of BiLSTM, bert-base-cased (BERT), and robert-base (RoBERTa) models on EnCBP-country and EnCBP-district. Average F1-macro scores over five runs with different random seeds are reported and standard deviations are shown in parentheses.}
\label{tbl:benchmark-performance}
\end{table}

After data validation, we divide both EnCBP-country and EnCBP-district into training, development, and test sets with a 80\%/10\%/10\% split and a random state of 42.
To show the predictability of cultural background labels with NLP models, we benchmark the EnCBP-country and EnCBP-district separately with BiLSTM, bert-base-cased, and roberta-base models.
We train the BiLSTM model for 20 epochs with a learning rate of 0.25 and fine-tune the other models for five epochs with a learning rate of 1e-4 on both subsets.

Table \ref{tbl:benchmark-performance} displays the average F1-macro scores across five runs with different random seeds for model initialization.
For all the models, the standard deviations of the five runs are at most 0.98 on EnCBP-country and 1.84 on EnCBP-district, indicating that randomness does not severely affect the predictions of models, and that the culture-specific writing styles can be modeled by DL models.
Both the BERT and RoBERTa models outperform the BiLSTM model with large margins, which suggests the importance of deep neural network architectures and large-scale pre-training for the task.
We also note that all the three models perform worse on EnCBP-district, which may be caused by both the more difficult task setting and the higher level of noise in EnCBP-district, since local news outlets target audiences from all over the country.
In the rest of this paper, we use the bert-base-cased model for the analyses and discussions since it is less resource-consuming than the roberta-base model, while the findings potentially apply to other model architectures as well.

\section{Cultural Domain Compatibility} \label{sct:dataset-analyses}

\begin{table*}[ht]
\centering
\begin{tabular}{|c|c|r|r|r|r|r|r|r|r|r|} 
\hline
\multicolumn{2}{|l|}{\multirow{2}{*}{}} & \multicolumn{9}{c|}{Evaluation Corpus}                                                                                                                                                                                                        \\ 
\cline{3-11}
\multicolumn{2}{|l|}{}                  & \multicolumn{1}{c|}{US} & \multicolumn{1}{c|}{UK} & \multicolumn{1}{c|}{AUS} & \multicolumn{1}{c|}{CAN} & \multicolumn{1}{c|}{IND} & \multicolumn{1}{c|}{NY} & \multicolumn{1}{c|}{CA} & \multicolumn{1}{c|}{GA} & \multicolumn{1}{c|}{TX}  \\ 
\hline
\multirow{9}{*}{\rotatebox[origin=c]{90}{Training Corpus}} & US   & \textbf{22.80}                   & 24.13                   & 25.08                    & 27.67                    & 26.54                    & \ul{28.08}                   & 24.54                   & 27.54                   & 24.41                    \\ 
\cline{2-11}
                                 & UK   & 24.77                   & \textbf{14.09}                   & 28.76                    & \ul{28.99}                    & 27.30                    & 25.50                   & 22.37                   & 26.30                   & 24.14                    \\ 
\cline{2-11}
                                 & AUS  & 22.49                   & \ul{27.56}                   & \textbf{21.82}                    & 26.53                    & 27.26                    & 25.31                   & 24.18                   & 23.69                   & 25.61                    \\ 
\cline{2-11}
                                 & CAN  & 26.13                   & \ul{37.45}                   & 30.60                    & \textbf{23.30}                    & 28.41                    & 24.32                  & 31.04                   & 26.30                   & 25.56                    \\ 
\cline{2-11}
                                 & IND  & 27.87                   & 24.63                   & 29.36                    & \ul{30.19}                    & \textbf{23.91}                    & 29.69                   & 26.46                   & 34.42                   & 26.40                    \\ 
\cline{2-11}
                                 & NY   & 22.65                   & 22.98                   & 25.68                    & 21.82                    & 25.66                    & \textbf{20.53}                   & 21.22                   & 22.98                   & \ul{25.88}                    \\ 
\cline{2-11}
                                 & CA   & 24.23                   & \ul{29.50}                   & 25.53                    & 24.41                    & 24.45                    & 24.77                   & \textbf{23.80}                   & 28.27                   & 27.92                    \\ 
\cline{2-11}
                                 & GA   & \textbf{19.21}                   & 24.61                   & \ul{29.29}                    & 26.76                    & 27.16                    & 21.44                   & 22.78                   & 20.25                   & 20.97                    \\ 
\cline{2-11}
                                 & TX   & 24.99                   & 26.96                   & \ul{30.91}                    & 29.97                    & 30.09                    & 30.31                   & 27.46                   & 26.64                   & \textbf{23.83}                    \\
\hline
\end{tabular}
\caption{Perplexity of LMs fine-tuned on the training corpora of EnCBP with the MLM objective and evaluated on the test corpora. The lowest perplexity for each fine-tuned LM is in bold and the highest perplexity is underlined.}
\label{tbl:domain-consistency-label}
\end{table*}

This section examines whether linguistic expressions are clearly separable across culture groups in EnCBP through LM evaluations.
We also manually examine representative linguistic expressions associated with each label to illustrate the differences in linguistic expression across cultures.
\subsection{Language Modeling Analysis}
Since all the documents in EnCBP come from news articles, we assume they are well-written and grammatically correct.
In addition, LMs trained on a grammatical corpus should produce similar perplexities on the corpus with each label if the writing styles are consistent across corpora.
Thus, to examine culture-specific differences in writing styles, we fine-tune a bert-base-cased model on the training corpus of each class in EnCBP with the MLM objective and evaluate perplexity of the fine-tuned models on all the test corpora.

As Table \ref{tbl:domain-consistency-label} shows, BERT models usually produce the lowest perplexities on the test portions of their training corpora, and the cross-corpus perplexities are usually considerably higher.
This supports our hypothesis that English writing styles are culture-dependent, and that the writing styles across cultures are different enough to be detected by LMs.
Meanwhile, we find that the cultural domain compatibility differs for different pairs of corpora, e.g., the IND corpus is more compatible with the UK corpus than other countries or districts.
The relations are not symmetric either, e.g., while the LM trained on the CAN corpus well adapts to the US corpus, the US LM performs the worst on the CAN corpus among the five countries.
These potentially result from the effects of geographical, geo-political, and historical backgrounds on the formation of cultural backgrounds. 
For instance, the US could be said to have greater influence on Canadian culture than vice versa. 
Potentially for similar reasons, compared to TX and GA, NY has a more consistent writing style with CAN.
We also note clear cultural domain compatibility gaps between liberal (NY and CA) and conservative states (GA and TX), which, agreeing with \citet{culture-classify-ideology-1}, shows that ideologies and policies of a district potentially has an effect on its culture-specific writing styles.
We provide additional topic-level LM analysis in Appendix \ref{sct:lm-analysis-additional}.


\subsection{Topic and Sentiment Distributions}
To verify if the different expressions across classes in the EnCBP datasets are triggered by cultural differences, we analyze the distributions of topics and sentiment scores for each class.
Specifically, we model the topics of each corpus using BERTopic \cite{bertopic-orig} and analyze sentiments of text using Stanza \cite{stanza-orig}.

We apply two-sided Kolmogorov-Smirnov (KS) tests on the topic distributions of each pair of classes to see whether the topic distributions for each country or state are similar. 
For all pairwise comparisons, the null hypothesis (which is that the distributions are identical) cannot be rejected using the KS test, with all p-values being above 0.1, and most in fact being above 0.7.
This potentially results from both topic control at the data collection phase and data filtering eliminating paragraphs containing NEs with high IDF scores.
Additionally, the sentiment score distribution is relatively consistent across classes (28.02\% to 34.97\% instances with negative sentiments).
Since the classes in EnCBP contain documents that are similar in topics and sentiments, it is likely that the differences in linguistic expressions across classes are caused by cultural differences.

\subsection{Manual Analysis} \label{sct:manual-analysis}
In addition to automatic evaluations, we manually examine distinguishable English expressions for each culture group in EnCBP.
Specifically, we extract phrases with high TF-IDF values for each corpus in EnCBP, retrieve news paragraphs that contain these phrases, and examine sentence structures and phrase usages in these representative instances.

From our analyses, we find that the different writing styles of countries and districts in EnCBP are affected by the choice of words or phrases, the ordering of phrases, and degrees of formality.
For example, the phrases ``in the wake of", ``in the lead up to", and ``the rest of the world" are much more frequently used by AUS news outlets than the others.
Also, the use of auxiliaries, especially the word ``may", is more frequent in the UK corpus, in the context of politeness.
The US corpus is in general more colloquial than the other corpora, as the journalists often write subjective comments in the news articles.
Additionally, the ways of referencing speeches differ across corpora, e.g., the quoted text usually appears prior to the ``[name] said" in the UK corpus but reversely in the US corpus.
In the EnCBP-district subset, the sentence structures are more consistent across corpora, while the mentions of NEs and wordings differ more.
For example, the word ``border" appears frequently in the TX corpus but less in the other corpora when discussing the ``immigration" topic.
Though the observations summarized from EnCBP may not be universally applicable to other datasets or text domains, they are validated by native speakers of English to be accounting for the high ACC in manual validations.

\section{Experiments and Analyses}
Since cultural background labels are expensive to annotate, most NLP models forego the use of this information to opt for larger training data amount.
For example, BERT is trained on Wikipedia text written in styles from mixed cultural backgrounds without access to cultural background information of the writers.
Using the EnCBP dataset we constructed, this section examines the relatedness between the cultural background prediction task and multiple other NLP tasks via model probing.
We also examine the effectiveness of cultural feature augmentation, i.e., augmenting DL models on downstream NLP tasks with culture-specific writing style information.
Specifically, we evaluate a bert-base-cased model with two common information injection methods, namely two-stage training and MTL, on nine syntactic, semantic, and psycholinguistic tasks.

\subsection{Tasks and Datasets}
The datasets used in our evaluations are:
\\
\textbf{PAWS-Wiki} \cite{PAWS-Wiki-orig} is a PI dataset containing English Wikipedia articles.
Each instance in PAWS-Wiki consists of a pair of sentences and a label indicating whether the two sentences are paraphrase (1) or not (0).
There are 49,401 training instances, 8,000 development instances, and 8,000 test instances in this dataset.
\\
\textbf{CoNLL-2003} English NER dataset \cite{conll-2003-orig} contains news articles from Reuters news only, so the dataset has a more consistent UK writing style, compared to the other datasets we utilize.
Each word in the documents is annotated with persons (PER), organizations (ORG), locations (LOC), or miscellaneous names (MISC) NE label in the IOB-2 format.
We adopt the official data split of the CoNLL-2003 dataset in the experiments, where there are 7,140, 1,837, and 1,668 NEs in the training, development, and test sets, respectively.
\\
\textbf{Go-Emotions} \cite{go-emotions-orig} is an ER dataset containing 58,009 English Reddit comments.
Instances in this dataset are labeled with 28 emotion types including neutral, in the multi-label classification form.
We split the dataset into training, development, and test sets with a 80\%/10\%/10\% split using 42 as the random seed.
To be consistent with other evaluations, we switch the annotations to the multi-class classification form by duplicating the data points associated with multiple labels and assigning one emotion label to each copy.
This results in an ER dataset containing 199,461 training instances, 35,057 development instances, and 34,939 test instances after removing instances with no labels.
\\
\textbf{Stanford Sentiment Treebank (SST-5)} \cite{SST-orig} is a document-level SA dataset containing sentences from movie reviews.
The documents are annotated with sentiment scores, which are turned to fine-grained (5-class) sentiment labels after pre-processing.
Using the official data split, we divide the dataset into training, development, and test splits containing 156,817, 1,102, and 2,211 instances, respectively.
Note that the training set of SST-5 contains a mixture of phrases and sentences, while the development and test sets contain only complete sentences.
\\
\textbf{SST-2} is the coarse-grained SST-5 dataset, in which each document is labeled with positive (1) or negative (0) sentiments.
There are 67,349 training instances, 872 development instances, and 1,821 test instances in this dataset.
\\
\textbf{QNLI} \cite{glue-orig} is a natural language inference (NLI) dataset with a question answering background.
Each instance in QNLI contains a question, a statement, and a label indicating whether the statement contains the answer to the question (1) or not (0).
There are 104,743 training instances, 5,463 development instances, and 5,463 test instances in this dataset.
\\
\textbf{STS-B} \cite{stsb-orig} is a benchmarked semantic textual similarity (STS) dataset.
Each instance in STS-B is a pair of sentences manually annotated with a semantic similarity score from 0 to 5.
The dataset contains 5,749 training instances, 1,500 development instances, and 1,379 test instances.
\\
\textbf{RTE} is a textual entailment (TE) dataset.
Each instance in RTE contains a pair of sentences and a label indicating whether the second sentence is an entailment (1) or not (0) of the first sentence.
The RTE dataset we use is a combination of RTE1 \cite{rte-1}, RTE2 \cite{rte-2}, RTE3 \cite{rte-3}, and RTE5 \cite{rte-5} datasets, which contains 2,490 training instances, 277 development instances, and 3,000 test instances.
\\
\textbf{Emotion} \cite{emotion-orig} is a Twitter-based ER dataset labeled with six emotion types, i.e., sadness (0), joy (1), love (2), anger (3), fear (4), and surprise (5).
There are 16,000 training instances, 2,000 development instances, and 2,000 test instances in this dataset.

\begin{table*}[!h]
\centering
\begin{tabular}{lcccc}
\hline
                      & \multicolumn{1}{c}{PAWS-Wiki (PI)} & \multicolumn{1}{c}{CoNLL-2003 (NER)} & \multicolumn{1}{c}{Go-Emotions (ER)} & \multicolumn{1}{c}{SST-5 (SA)} \\ \hline
BERT-orig                  &90.01 (0.35)&91.73 (0.39)&31.67 (0.59)&52.41 (1.20)\\ \hline
+ two-stage training  &91.67 (0.20)&94.41 (0.10)&30.72 (0.16)&54.54 (0.45)\\ \hline
+ multi-task learning &91.58 (0.19)&92.92 (0.18)&30.71 (0.24)&54.47 (0.70)\\ \hline
\end{tabular}
\setlength{\tabcolsep}{6.5pt}
\begin{tabular}{lccccc}

                      & QNLI (NLI)         & STS-B (STS)                                                            & RTE (TE)         & SST-2 (SA)        & Emotion (ER)      \\ \hline
BERT-orig             & 90.89 (0.06) & \begin{tabular}[c]{@{}c@{}}89.22/88.83\\ (0.05/0.02)\end{tabular} & 64.69 (1.13) & 91.86 (0.46) & 88.25 (0.49) \\ \hline
+ two-stage training  & 91.77 (0.09) & \begin{tabular}[c]{@{}c@{}}89.47/89.08\\ (0.11/0.13)\end{tabular} & 68.45 (1.71) & 93.09 (0.33) & 91.94 (0.50) \\ \hline
+ multi-task learning & 91.20 (0.22) & \begin{tabular}[c]{@{}c@{}}89.32/88.94\\ (0.10/0.11)\end{tabular} & 70.76 (0.93) & 92.34 (0.42) & 91.70 (0.35) \\ \hline
\end{tabular}
\caption{The performance of BERT model without cultural feature augmentation (BERT-orig), and models with cultural feature augmentation via two-stage training and multi-task learning. \textbf{EnCBP-country} is used as the auxiliary dataset. We report accuracy for QNLI, RTE, and SST-2, Pearson’s and Spearman’s correlations for STS-B, and F1-macro for the other tasks. The average score and standard deviation (in parentheses) in five runs with different random seeds are reported for each experiment}
\label{tbl:two-stage-mtl-coarse-grained}
\end{table*}

\subsection{Feature Augmentation} \label{sct:feature-augmentation}
\subsubsection{Experimental Settings}
We use the Huggingface \cite{huggingface-orig} implementation of BERT in all our evaluations.
On each task, we fine-tune a bert-base-cased model for five epochs with different random seeds, and we report the average evaluation score on the test sets of downstream tasks over the five runs to avoid the influence of randomness.
Each experiment is run on a single RTX-6000 GPU with a learning rate of 1e-4 and a batch size of 32.

\subsubsection{Two-Stage Training}
\begin{table*}[!h]
\centering
\begin{tabular}{lcccc}
\hline
                      & \multicolumn{1}{c}{PAWS-Wiki (PI)} & \multicolumn{1}{c}{CoNLL-2003 (NER)} & \multicolumn{1}{c}{Go-Emotions (ER)} & \multicolumn{1}{c}{SST-5 (SA)} \\ \hline
BERT-orig                  &90.01 (0.35)&91.73 (0.39)&31.67 (0.59)&52.41 (1.20)\\ \hline
+ two-stage training  &91.40 (0.20)&94.25 (0.11)&30.21 (0.37)&53.82 (0.45)\\ \hline
+ multi-task learning &91.70 (0.23)&93.64 (0.14)&30.47 (0.14)&53.52 (0.54)\\ \hline
\end{tabular}
\setlength{\tabcolsep}{6.5pt}
\begin{tabular}{lccccc}
                      & QNLI (NLI)         & STS-B (STS)                                                            & RTE (TE)         & SST-2 (SA)        & Emotion (ER)      \\ \hline
BERT-orig             & 90.89 (0.06) & \begin{tabular}[c]{@{}c@{}}89.22/88.83\\ (0.05/0.02)\end{tabular} & 64.69 (1.13) & 91.86 (0.46) & 88.25 (0.49) \\ \hline
+ two-stage training  & 91.77 (0.08) & \begin{tabular}[c]{@{}c@{}}89.45/89.01\\ (0.12/0.13)\end{tabular} & 67.87 (1.09) & 92.52 (0.32) & 91.65 (0.24) \\ \hline
+ multi-task learning & 91.21 (0.24) & \begin{tabular}[c]{@{}c@{}}89.34/89.14\\ (0.11/0.10)\end{tabular} & 69.68 (1.04) & 92.89 (0.36) & 92.07 (0.52) \\ \hline
\end{tabular}
\caption{The performance of BERT model without cultural feature augmentation (BERT-orig), and models with cultural feature augmentation via two-stage training and multi-task learning. The \textbf{EnCBP-district} is used as the auxiliary dataset. We report accuracy for QNLI, RTE, and SST-2, Pearson’s and Spearman’s correlations for STS-B, and F1-macro for the other tasks. The average score and standard deviation (in parentheses) in five runs with different random seeds are reported for each experiment}
\label{tbl:two-stage-mtl-finer-grained}
\end{table*}
We first explore the two-stage training method which successively fine-tunes the pre-trained BERT model on a cultural background prediction dataset and the target task.
We use EnCBP-country here to examine the efficacy of coarse-grained cultural feature augmentation, and we study the effect of using EnCBP-district in Section \ref{sct:experiments-fine-grained}.

As Table \ref{tbl:two-stage-mtl-coarse-grained} shows, the two-stage training strategy brings noticeable performance improvements to the SA models.
This agrees with prior psychological research \cite{culture-emotion-2}, since the expressions of sentiments and attitudes differ across culture groups.
Similarly, since NEs are usually mentioned differently across cultures, training the model to distinguish culture-specific writing styles helps resolve the conflict between the training domain of BERT and that of the CoNLL-2003 dataset and improves the performance of the NER model.
On the PI task, while two-stage training has a positive effect on the performance of the model, the score improvement is not as significant as those on SA and NER tasks.
The same trend holds for two other semantic tasks (QNLI and STS-B), where two-stage training brings only marginal performance improvements.
We attribute this to the additional noise introduced by the cultural background labels for a semantic task, since expressions with the same semantic meaning can be associated with different cultural background labels in EnCBP.
To verify this assumption, we conduct an additional experiment by applying the MLM objective instead of the classification objective in the first training stage.
The model performance on PI is raised to 94.11 in F1-macro score, outperforming the previous two-stage training model by 2.44.
The two-stage training performance also improves by 0.81 and 0.49/0.53 for QNLI and STS-B when using the MLM objective at the first fine-tuning stage.
These results imply that while the cultural background labels are noisy for semantic tasks, enhancing the LM with English expressions from multiple cultural backgrounds is beneficial.
Quite differently, however, two-stage training brings noticeable performance improvements to the RTE model.
One possible explanation is, as is supported by the large standard deviations of evaluation scores in five runs, that the RTE dataset is too small and the performance tend to be affected more greatly by other issues such as model initialization.
Unlike the other tasks, the performance of BERT drops on Go-Emotions in our evaluations, which is counter-intuitive since expressions of emotion are culture-specific \cite{culture-emotion-1}.
We hypothesize that the negative effect of cultural feature augmentation is mainly caused by the imbalanced distribution of users' cultural backgrounds in the Go-Emotions dataset, as the dataset is constructed over a Reddit \footnote{https://www.reddit.com/} corpus and nearly 50\% Reddit users are from the US \footnote{https://www.statista.com/statistics/325144/reddit-global-active-user-distribution/}.
Supporting our hypothesis, cultural feature augmentation on the Emotion dataset notably improves the performance of BERT, despite the domain differences between the EnCBP-country (news domain) and Emotion (social media domain) datasets.

\subsubsection{Multi-Task Learning}
We further explore MTL methods for cultural feature augmentation when training the BERT model on downstream tasks.
Specifically, we use EnCBP-country as the auxiliary task and train the model alternatively on the primary and auxiliary tasks.

According to Table \ref{tbl:two-stage-mtl-coarse-grained}, introducing cultural background information via MTL improves the performance of BERT on all the datasets except for Go-Emotions, similar to the two-stage training method.
However, the performance on NER is noticeably lower with MTL than with two-stage training.
This potentially results from the mono-cultural nature of the CoNLL-2003 dataset, which is constructed on Reuters news, a UK news outlet.
While the information and expressions in countries other than UK fade gradually during the second training stage, the MTL method strengthens the irrelevant information in the entire training process and harms the evaluation performance of the model more severely.
To validate our hypothesis, we generate a binary cultural background prediction dataset by treating the UK documents as positive instances and the others as negative instances, and we re-run the MTL evaluation on the CoNLL-2003 dataset.
The performance of BERT under this setting is raised to 93.97 in F1-macro score, which implies the importance of careful text domain selection for improving the performance of DL models with cultural feature augmentation.

\subsubsection{Finer-Grained Feature Augmentation} \label{sct:experiments-fine-grained}
We repeat the two-stage training and MTL evaluations on the nine downstream tasks using EnCBP-district to examine the effects of cultural feature augmentation with cultural background information with different granularity levels.
The evaluation results are shown in Table \ref{tbl:two-stage-mtl-finer-grained}.
While the scores are very consistent with those in Table \ref{tbl:two-stage-mtl-coarse-grained}, we observe better MTL performance on CoNLL-2003 and Emotion and worse performance with both two-stage training and MTL on SST-5.
Based on our analysis of EnCBP-country and EnCBP-district, the larger gaps in writing style among countries than those across states are likely the cause of the lower NER evaluation performance.
In EnCBP-district, the linguistic expressions are more consistent since they all come from news outlets in the US, which relieves the problem and improves the MTL performance on CoNLL-2003.
On the contrary, the lower diversity in expressions potentially negatively affects the performance of the SST-5 model since the SA task benefits from identifying culture-specific linguistic expressions, and since the corpus of SST-5 contains writings from all over the world.
In addition, using EnCBP-district does not relieve the problem on the Go-Emotions dataset either, which suggests the limitation of cultural feature augmentation: trying to distinct expressions in different cultural backgrounds may introduce unexpected noise into models especially when the cultural background of a dataset is mostly the same.
The performance of BERT on the Emotion dataset which consists of writings from more diverse cultural backgrounds, for example, is subject to comparable or even greater improvements when the model is augmented using the finer-grained EnCBP-district dataset.

To summarize, while cultural feature augmentation using EnCBP is beneficial for a wide range of NLP tasks, the necessity of conducting cultural feature augmentation has to be carefully evaluated.
We also examine the effect of feature augmentation with less auxiliary data in Appendix \ref{appendix:less-data}, showing that the size of the auxiliary data has an affect on the performance of DL models.


\section{Conclusion and Future Work}
This paper presents EnCBP, a mono-lingual news-based cultural background prediction dataset containing country-level (coarse-grained) and district-level (finer-grained) cultural background labels.
Through manual validation on MTurk and cultural domain compatibility evaluations, we find that writing style clearly differs across countries and districts, confirming that cultural background has a substantial effect on writing style even in the same language.
We also benchmark the dataset with state-of-the-art NLP models to show that, though challenging, different English expressions across cultural backgrounds can be identified and classified into culture categories by DL models.
Additionally, our evaluations on downstream NLP tasks of various types show that cultural feature augmentation is able to improve the performance of DL models on various semantic, syntactic, and psycholinguistic tasks.
While the performance of the BERT model is negatively affected by introducing cultural background information on an ER dataset, the imbalanced distribution of cultural backgrounds in its corpus may account for the performance drop.
Our results demonstrate that cultural feature augmentation with EnCBP is a practical way of improving the performance of DL models on various NLP tasks, as long as the text domains of EnCBP and the downstream tasks are not too divergent.

Future work can extend our research to examine cultural differences in social media writings, which reflect even finer-grained cultural distinctions and are much noisier and difficult to annotate or validate than news articles.

\section{Ethics Statement and Broader Impacts}
This paper presents and releases a news-based cultural background prediction dataset.
The dataset is constructed on publicly available news outlets using the public API of Media Cloud and the labels are generated based on the country and district codes of the media outlets.
Thus, there is no sensitive or private information in the dataset.
Additionally, since we use mainstream news outlets for our data collection we believe there is less risk of overtly unethical information (though we cannot be sure given the current sociopolitical climate). Given the relatively large size of our dataset, we cannot manually examine all articles, however, the publicly released dataset will warn users of the possibility of the dataset containing unethical information and will allows users to flag unethical articles in our dataset.
We also hired annotators from MTurk to validate the quality of annotations for a sample instances from our dataset.
To ensure the quality of dataset validation, we require the annotators to be native English speakers from the same country or district as the label of each instance to be validated.
The annotators were given clear instructions to choose the news paragraph(s) written by journalists in their countries or districts from a pair of paragraphs.
We payed \$0.14 (USD) for validating each instance, which translates to over \$25 per hour since each data point takes no more than 1 minute to validate.
This hourly rate is considerably higher than the federal minimum wage in the US.
The entire annotation process was anonymized and the annotators were not asked for their personally identifiable information, so there was not any risk of harm associated with their participation.

This paper presents one of the first attempts at tailoring NLP models to the writing styles of specific regions, thus reducing the out-sized influence of the linguistic style of larger countries in these models.

\bibliography{anthology,custom}

\begin{thebibliography}{28}
\expandafter\ifx\csname natexlab\endcsname\relax\def\natexlab#1{#1}\fi

\bibitem[{Acheampong et~al.(2020)Acheampong, Wenyu, and
  Nunoo-Mensah}]{culture-emotion-3}
Francisca~Adoma Acheampong, Chen Wenyu, and Henry Nunoo-Mensah. 2020.
\newblock Text-based emotion detection: Advances, challenges, and
  opportunities.
\newblock \emph{Engineering Reports}, 2(7):e12189.

\bibitem[{Almuhailib(2019)}]{culture-writing-style-2}
Badar Almuhailib. 2019.
\newblock Analyzing cross-cultural writing differences using contrastive
  rhetoric: A critical review.
\newblock \emph{Advances in Language and Literary Studies}, 10(2):102--106.

\bibitem[{Bar-Haim et~al.(2006)Bar-Haim, Dagan, Dolan, Ferro, Giampiccolo,
  Magnini, and Szpektor}]{rte-2}
Roy Bar-Haim, Ido Dagan, Bill Dolan, Lisa Ferro, Danilo Giampiccolo, Bernardo
  Magnini, and Idan Szpektor. 2006.
\newblock The second pascal recognising textual entailment challenge.
\newblock In \emph{Proceedings of the second PASCAL challenges workshop on
  recognising textual entailment}, volume~6, pages 6--4. Venice.

\bibitem[{Bentivogli et~al.(2009)Bentivogli, Clark, Dagan, and
  Giampiccolo}]{rte-5}
Luisa Bentivogli, Peter Clark, Ido Dagan, and Danilo Giampiccolo. 2009.
\newblock The fifth pascal recognizing textual entailment challenge.
\newblock In \emph{TAC}.

\bibitem[{Cer et~al.(2017)Cer, Diab, Agirre, Lopez-Gazpio, and
  Specia}]{stsb-orig}
Daniel Cer, Mona Diab, Eneko Agirre, Inigo Lopez-Gazpio, and Lucia Specia.
  2017.
\newblock Semeval-2017 task 1: Semantic textual similarity-multilingual and
  cross-lingual focused evaluation.
\newblock \emph{arXiv preprint arXiv:1708.00055}.

\bibitem[{Dagan et~al.(2005)Dagan, Glickman, and Magnini}]{rte-1}
Ido Dagan, Oren Glickman, and Bernardo Magnini. 2005.
\newblock The pascal recognising textual entailment challenge.
\newblock In \emph{Machine Learning Challenges Workshop}, pages 177--190.
  Springer.

\bibitem[{Demszky et~al.(2020)Demszky, Movshovitz-Attias, Ko, Cowen, Nemade,
  and Ravi}]{go-emotions-orig}
Dorottya Demszky, Dana Movshovitz-Attias, Jeongwoo Ko, Alan Cowen, Gaurav
  Nemade, and Sujith Ravi. 2020.
\newblock \href {https://doi.org/10.18653/v1/2020.acl-main.372}
  {{G}o{E}motions: A dataset of fine-grained emotions}.
\newblock In \emph{Proceedings of the 58th Annual Meeting of the Association
  for Computational Linguistics}, pages 4040--4054, Online. Association for
  Computational Linguistics.

\bibitem[{Fleiss(1971)}]{fleiss-kappa-orig}
Joseph~L Fleiss. 1971.
\newblock Measuring nominal scale agreement among many raters.
\newblock \emph{Psychological bulletin}, 76(5):378.

\bibitem[{Giampiccolo et~al.(2007)Giampiccolo, Magnini, Dagan, and
  Dolan}]{rte-3}
Danilo Giampiccolo, Bernardo Magnini, Ido Dagan, and Bill Dolan. 2007.
\newblock The third pascal recognizing textual entailment challenge.
\newblock In \emph{Proceedings of the ACL-PASCAL workshop on textual entailment
  and paraphrasing}, pages 1--9. Association for Computational Linguistics.

\bibitem[{Grootendorst(2020)}]{bertopic-orig}
Maarten Grootendorst. 2020.
\newblock \href {https://doi.org/10.5281/zenodo.4381785} {Bertopic: Leveraging
  bert and c-tf-idf to create easily interpretable topics.}

\bibitem[{Hareli et~al.(2015)Hareli, Kafetsios, and Hess}]{culture-emotion-1}
Shlomo Hareli, Konstantinos Kafetsios, and Ursula Hess. 2015.
\newblock A cross-cultural study on emotion expression and the learning of
  social norms.
\newblock \emph{Frontiers in psychology}, 6:1501.

\bibitem[{Imran et~al.(2020)Imran, Doudpota, Kastrati, and
  Bhatra}]{culture-classify-ideology-1}
Ali~Shariq Imran, Sher~Mohammad Doudpota, Zenun Kastrati, and Rakhi Bhatra.
  2020.
\newblock Cross-cultural polarity and emotion detection using sentiment
  analysis and deep learning--a case study on covid-19.
\newblock \emph{arXiv preprint arXiv:2008.10031}.

\bibitem[{Kim(2002)}]{culture-talk-1}
Heejung~S Kim. 2002.
\newblock We talk, therefore we think? a cultural analysis of the effect of
  talking on thinking.
\newblock \emph{Journal of personality and social psychology}, 83(4):828.

\bibitem[{Kitano(1990)}]{culture-writing-style-3}
Hiroko Kitano. 1990.
\newblock Cross-cultural differences in written discourse patterns: a study of
  acceptability of japanese expository compositions in american universities.

\bibitem[{Krampetz(2005)}]{culture-writing-style-1}
Erin~McClure Krampetz. 2005.
\newblock International educational administration and policy analysis.

\bibitem[{Liu et~al.(2019)Liu, Ott, Goyal, Du, Joshi, Chen, Levy, Lewis,
  Zettlemoyer, and Stoyanov}]{roberta-orig}
Yinhan Liu, Myle Ott, Naman Goyal, Jingfei Du, Mandar Joshi, Danqi Chen, Omer
  Levy, Mike Lewis, Luke Zettlemoyer, and Veselin Stoyanov. 2019.
\newblock Roberta: A robustly optimized bert pretraining approach.
\newblock \emph{arXiv preprint arXiv:1907.11692}.

\bibitem[{Nisbett et~al.(2001)Nisbett, Peng, Choi, and
  Norenzayan}]{culture-think-1}
Richard~E Nisbett, Kaiping Peng, Incheol Choi, and Ara Norenzayan. 2001.
\newblock Culture and systems of thought: holistic versus analytic cognition.
\newblock \emph{Psychological review}, 108(2):291.

\bibitem[{Qi et~al.(2020)Qi, Zhang, Zhang, Bolton, and Manning}]{stanza-orig}
Peng Qi, Yuhao Zhang, Yuhui Zhang, Jason Bolton, and Christopher~D. Manning.
  2020.
\newblock \href {https://nlp.stanford.edu/pubs/qi2020stanza.pdf} {Stanza: A
  {Python} natural language processing toolkit for many human languages}.
\newblock In \emph{Proceedings of the 58th Annual Meeting of the Association
  for Computational Linguistics: System Demonstrations}.

\bibitem[{Saravia et~al.(2018)Saravia, Liu, Huang, Wu, and Chen}]{emotion-orig}
Elvis Saravia, Hsien-Chi~Toby Liu, Yen-Hao Huang, Junlin Wu, and Yi-Shin Chen.
  2018.
\newblock \href {https://doi.org/10.18653/v1/D18-1404} {{CARER}: Contextualized
  affect representations for emotion recognition}.
\newblock In \emph{Proceedings of the 2018 Conference on Empirical Methods in
  Natural Language Processing}, pages 3687--3697, Brussels, Belgium.
  Association for Computational Linguistics.

\bibitem[{Socher et~al.(2013)Socher, Perelygin, Wu, Chuang, Manning, Ng, and
  Potts}]{SST-orig}
Richard Socher, Alex Perelygin, Jean Wu, Jason Chuang, Christopher~D. Manning,
  Andrew Ng, and Christopher Potts. 2013.
\newblock \href {https://www.aclweb.org/anthology/D13-1170} {Recursive deep
  models for semantic compositionality over a sentiment treebank}.
\newblock In \emph{Proceedings of the 2013 Conference on Empirical Methods in
  Natural Language Processing}, pages 1631--1642, Seattle, Washington, USA.
  Association for Computational Linguistics.

\bibitem[{Sun et~al.(2021)Sun, Ahn, Park, Tsvetkov, and
  Mortensen}]{culture-emotion-2}
Jimin Sun, Hwijeen Ahn, Chan~Young Park, Yulia Tsvetkov, and David~R.
  Mortensen. 2021.
\newblock \href {https://doi.org/10.18653/v1/2021.eacl-main.204}
  {Cross-cultural similarity features for cross-lingual transfer learning of
  pragmatically motivated tasks}.
\newblock In \emph{Proceedings of the 16th Conference of the European Chapter
  of the Association for Computational Linguistics: Main Volume}, pages
  2403--2414, Online. Association for Computational Linguistics.

\bibitem[{Tambassi(2018)}]{culture-specific-language-1}
Timothy Tambassi. 2018.
\newblock \href {https://doi.org/10.4000/estetica.2752} {From geographical
  lines to cultural boundaries: Mapping the ontological debate}.
\newblock \emph{Rivista di estetica}, 67:150--164.

\bibitem[{Tian et~al.(2021)Tian, Chakrabarty, Morstatter, and
  Peng}]{culture-classify-language-1}
Yufei Tian, Tuhin Chakrabarty, Fred Morstatter, and Nanyun Peng. 2021.
\newblock \href {https://doi.org/10.18653/v1/2021.socialnlp-1.16} {Identifying
  distributional perspectives from colingual groups}.
\newblock In \emph{Proceedings of the Ninth International Workshop on Natural
  Language Processing for Social Media}, pages 178--190, Online. Association
  for Computational Linguistics.

\bibitem[{Tjong Kim~Sang and De~Meulder(2003)}]{conll-2003-orig}
Erik~F. Tjong Kim~Sang and Fien De~Meulder. 2003.
\newblock \href {https://aclanthology.org/W03-0419} {Introduction to the
  {C}o{NLL}-2003 shared task: Language-independent named entity recognition}.
\newblock In \emph{Proceedings of the Seventh Conference on Natural Language
  Learning at {HLT}-{NAACL} 2003}, pages 142--147.

\bibitem[{Vaswani et~al.(2017)Vaswani, Shazeer, Parmar, Uszkoreit, Jones,
  Gomez, Kaiser, and Polosukhin}]{bert-orig}
Ashish Vaswani, Noam Shazeer, Niki Parmar, Jakob Uszkoreit, Llion Jones,
  Aidan~N. Gomez, Lukasz Kaiser, and Illia Polosukhin. 2017.
\newblock \href
  {https://proceedings.neurips.cc/paper/2017/hash/3f5ee243547dee91fbd053c1c4a845aa-Abstract.html}
  {Attention is all you need}.
\newblock In \emph{Advances in Neural Information Processing Systems 30: Annual
  Conference on Neural Information Processing Systems 2017, December 4-9, 2017,
  Long Beach, CA, {USA}}, pages 5998--6008.

\bibitem[{Wang et~al.(2019)Wang, Singh, Michael, Hill, Levy, and
  Bowman}]{glue-orig}
Alex Wang, Amanpreet Singh, Julian Michael, Felix Hill, Omer Levy, and
  Samuel~R. Bowman. 2019.
\newblock {GLUE}: A multi-task benchmark and analysis platform for natural
  language understanding.
\newblock In the Proceedings of ICLR.

\bibitem[{Wolf et~al.(2020)Wolf, Debut, Sanh, Chaumond, Delangue, Moi, Cistac,
  Rault, Louf, Funtowicz, Davison, Shleifer, von Platen, Ma, Jernite, Plu, Xu,
  Scao, Gugger, Drame, Lhoest, and Rush}]{huggingface-orig}
Thomas Wolf, Lysandre Debut, Victor Sanh, Julien Chaumond, Clement Delangue,
  Anthony Moi, Pierric Cistac, Tim Rault, Rémi Louf, Morgan Funtowicz, Joe
  Davison, Sam Shleifer, Patrick von Platen, Clara Ma, Yacine Jernite, Julien
  Plu, Canwen Xu, Teven~Le Scao, Sylvain Gugger, Mariama Drame, Quentin Lhoest,
  and Alexander~M. Rush. 2020.
\newblock \href {https://www.aclweb.org/anthology/2020.emnlp-demos.6}
  {Transformers: State-of-the-art natural language processing}.
\newblock In \emph{Proceedings of the 2020 Conference on Empirical Methods in
  Natural Language Processing: System Demonstrations}, pages 38--45, Online.
  Association for Computational Linguistics.

\bibitem[{Zhang et~al.(2019)Zhang, Baldridge, and He}]{PAWS-Wiki-orig}
Yuan Zhang, Jason Baldridge, and Luheng He. 2019.
\newblock {PAWS: Paraphrase Adversaries from Word Scrambling}.
\newblock In \emph{Proc. of NAACL}.

\end{thebibliography}
\bibliographystyle{acl_natbib}

\cleardoublepage
\appendix
\setcounter{table}{0}
\renewcommand{\thetable}{A\arabic{table}}

\section{Language Modeling Analysis Based on Topic} \label{sct:lm-analysis-additional}
\begin{table*}[h]
\centering
\begin{tabular}{|c|c|c|c|c|c|c|} 
\hline
\multicolumn{2}{|l|}{\multirow{2}{*}{}}                                                           & \multicolumn{5}{c|}{Evaluation Corpus}                                                                                                                                                                                                                                                                                  \\ 
\cline{3-7}
\multicolumn{2}{|l|}{}                                                                            & \multicolumn{1}{c|}{\begin{tabular}[c]{@{}c@{}}Global \\Warming\end{tabular}} & \multicolumn{1}{c|}{Abortion} & \multicolumn{1}{c|}{Immigration} & \multicolumn{1}{c|}{\begin{tabular}[c]{@{}c@{}}Social Safety\\Net\end{tabular}} & \multicolumn{1}{c|}{\begin{tabular}[c]{@{}c@{}}Mandatory \\Vaccines\end{tabular}}  \\ 
\hline
\multirow{5}{*}{\rotatebox[origin=c]{90}{Training Corpus}} & \begin{tabular}[c]{@{}c@{}}Global \\Warming\end{tabular}     & \textbf{21.42}                                                                         & 25.79                         & 25.29                            & \ul{26.36}                                                                           & 24.18                                                                              \\ 
\cline{2-7}
                                  & Abortion                                                     & 26.40                                                                         & \textbf{20.79}                         & \ul{30.66}                            & 24.38                                                                           & 25.80                                                                              \\ 
\cline{2-7}
                                  & Immigration                                                  & \ul{30.00}                                                                         & 25.00                         & 28.70                            & 25.50                                                                           & \textbf{24.88}                                                                              \\ 
\cline{2-7}
                                  & \begin{tabular}[c]{@{}c@{}}Social Safety\\Net\end{tabular}   & \textbf{25.54}                                                                         & 26.80                         & 27.78                            & \ul{29.01}                                                                           & 27.88                                                                              \\ 
\cline{2-7}
                                  & \begin{tabular}[c]{@{}c@{}}Mandatory \\Vaccines\end{tabular} & 25.48                                                                         & 25.13                         & \ul{29.53}                            & 28.18                                                                           & \textbf{23.22}                                                                             \\
\hline
\end{tabular}
\caption{Perplexity of each BERT model fine-tuned on a training topic with the MLM objective and evaluated on an evaluation topic. The lowest perplexity for each fine-tuned LM is in bold and the highest perplexity is underlined.}
\label{tbl:domain-consistency-topic}
\end{table*}
We study the cultural domain compatibility across news topics in EnCBP by repeating the LM evaluations with the news paragraphs grouped by their topics.
As Table \ref{tbl:domain-consistency-topic} shows, for the topics ``Immigration" and ``Social Safety Net", the LMs do not achieve the lowest perplexities on their training topics.
We speculate that this reflects the more controversial nature of these two topics, since linguistic expressions are heavily affected by attitudes and stances.
In addition, since each country or state news outlet has a relatively stable attitude towards each topic, the discrepancy between each trained LM and the test set in the cultural domain of its training set implies that the EnCBP dataset is constructed over diverse culture groups.
The diverse writing styles in EnCBP make it appropriate for improving DL models on downstream tasks via cultural feature augmentation, since EnCBP does not bias extremely towards the writing styles of a single culture group.

\renewcommand{\thetable}{B\arabic{table}}
\setcounter{table}{0}

\begin{table*}[]
\centering
\begin{tabular}{llcccc} 
\hline
\multirow{2}{*}{\begin{tabular}[c]{@{}l@{}}\\DR\end{tabular}} &                       & PAWS-Wiki (PI) & CoNLL-2003 (NER) & Go-Emotions (ER) & SST-5 (SA)  \\ 
\cline{2-6}
                                                              & BERT-orig             & 90.01          & 91.73            & 31.67            & 52.41     \\ 
\hline
\multirow{2}{*}{\rotatebox[origin=c]{90}{80\%}}                                             & + two-stage training  & 91.24          & 94.07            & 29.76            & 53.86     \\ 
\cline{2-6}
                                                              & + multi-task learning & 91.50          & 93.88            & 29.42            & 54.37     \\ 
\hline
\multirow{2}{*}{\rotatebox[origin=c]{90}{60\%}}                                             & + two-stage training  & 90.60          & 92.50            & 28.98            & 50.54     \\ 
\cline{2-6}
                                                              & + multi-task learning & 90.84          & 92.00            & 28.97            & 51.24     \\ 
\hline
\multirow{2}{*}{\rotatebox[origin=c]{90}{20\%}}                                             & + two-stage training  & 90.15          & 91.75            & 28.84            & 50.04     \\ 
\cline{2-6}
                                                              & + multi-task learning & 90.23          & 91.53            & 28.81            & 50.71     \\
\hline
\end{tabular}
\caption{The performance of BERT without cultural feature augmentation (BERT-orig), and models with cultural feature augmentation via two-stage training (+two-stage training) and multi-task learning (+multi-task learning). The downsampled \textbf{EnCBP-country} datasets are used as auxiliary datasets. DR represents the percentile of remaining data.}
\label{tbl:feature-augmentation-eval-coarse-subsampled-appendix}
\end{table*}

\begin{table*}[]
\centering
\begin{tabular}{llcccc} 
\hline
\multirow{2}{*}{\begin{tabular}[c]{@{}l@{}}\\DR\end{tabular}} &                       & PAWS-Wiki (PI) & CoNLL-2003 (NER) & Go-Emotions (ER) & SST-5 (SA)  \\ 
\cline{2-6}
                                                              & BERT-orig             & 90.01          & 91.73            & 31.67            & 52.41     \\ 
\hline
\multirow{2}{*}{\rotatebox[origin=c]{90}{80\%}}                                             & + two-stage training  & 91.18          & 93.48            & 29.57            & 53.34     \\ 
\cline{2-6}
                                                              & + multi-task learning & 90.91          & 93.29            & 29.96            & 53.38     \\ 
\hline
\multirow{2}{*}{\rotatebox[origin=c]{90}{60\%}}                                             & + two-stage training  & 90.23          & 93.34            & 28.43            & 51.86     \\ 
\cline{2-6}
                                                              & + multi-task learning & 90.46          & 92.85            & 28.54            & 51.06     \\ 
\hline
\multirow{2}{*}{\rotatebox[origin=c]{90}{20\%}}                                             & + two-stage training  & 89.98          & 92.00            & 28.81            & 50.71     \\ 
\cline{2-6}
                                                              & + multi-task learning & 90.00          & 91.65            & 28.39            & 50.02     \\
\hline
\end{tabular}
\caption{The performance of BERT without cultural feature augmentation (BERT-orig), and models with cultural feature augmentation via two-stage training (+two-stage training) and multi-task learning (+multi-task learning). The downsampled \textbf{EnCBP-district} datasets are used as auxiliary datasets. DR represents the percentile of remaining data.}
\label{tbl:feature-augmentation-eval-finer-subsampled-appendix}
\end{table*}

\section{Feature Augmentation with Less Data} \label{appendix:less-data}
We repeat the joint modeling and two-stage training experiments on PAWS-Wiki, CoNLL-2003, Go-Emotions, and SST-5 datasets with randomly downsampled EnCBP-country and EnCBP-district training datasets to examine the effect of auxiliary data size.
Specifically, we randomly reduce 20\%, 40\%, and 80\% of training instances from EnCBP-country and EnCBP-district with a random seed of 42 and use the reduced datasets in the evaluations.
The experimental results are shown in Table \ref{tbl:feature-augmentation-eval-coarse-subsampled-appendix} (EnCBP-country) and Table \ref{tbl:feature-augmentation-eval-finer-subsampled-appendix} (EnCBP-district).

While removing 20\% of the training instances from EnCBP-country and EnCBP-district generally does not greatly affect the feature augmentation evaluation results, there is noticeable performance gap on all the tasks when over 40\% of the training instances are eliminated.
This may be due to the poorer predictability of cultural background labels from the much smaller training datasets, as the BERT performance drops greatly from 78.13 to 60.92 (on EnCBP-country) and from 72.09 to 60.03 (on EnCBP-district) when 40\% of the training data is removed (see Table \ref{tbl:benchmark-performance} for the original BERT performance results).
On the other hand, though using more training data from EnCBP has positive overall effects on the performance of feature-augmented models, the improvements become gradually smaller when the training data amount increases.

In brief, through these experiments we hypothesize that a cultural background prediction dataset of a moderate size such as EnCBP is sufficient for cultural feature augmentation. Even if datasets larger in size could potentially lead to better performance improvements, the gains are likely to be small compared to the effort required for constructing a larger dataset.

\end{document}